\documentclass[letterpaper, 10 pt, conference]{ieeeconf}
\IEEEoverridecommandlockouts
\usepackage{amsmath,amsfonts}
\usepackage{subfigure}
\usepackage{graphicx}
\usepackage{balance}
\usepackage[ruled,linesnumbered,boxed,noend]{algorithm2e}
\usepackage{booktabs}
\usepackage{threeparttable}
\usepackage{hyperref}
\usepackage{array}
\usepackage{textcomp}
\usepackage{stfloats}
\usepackage{url}
\usepackage{verbatim}
\usepackage{cite}
\usepackage{algpseudocode}
\usepackage{stfloats}
\usepackage{microtype}
\usepackage{multirow}

\allowdisplaybreaks

\begin{document}

\setlength{\abovedisplayskip}{4pt}
\setlength{\belowdisplayskip}{4pt}

\title{\bfseries \parbox{0.85\linewidth}{\centering HCLM: A Hierarchical Framework for Cooperative Loco-Manipulation with Dual Quadrupeds}}

\author{ Qixuan Li$^{1,*}$, Chen Le$^{1,*}$, Jincheng Yu$^{2,\dagger}$, Xinlei Chen$^{1,\dagger}$
\thanks{$^*$ \textbf{Equal Contribution.} $^\dagger$ \textbf{Corresponding Authors.}}
\thanks{$^1$ Shenzhen International Graduate School, Tsinghua University, Shenzhen, China. lqx23@mails.tsinghua.edu.cn,  le-c25@mails.tsinghua.edu.cn, chen.xinlei@sz.tsinghua.edu.cn.} 
\thanks{$^2$ Department of Electronic Engineering, and the Institute for Embodied Intelligence and Robotics, Tsinghua University, Beijing, China. yu-jc@mail.tsinghua.edu.cn. }
}

\maketitle

\begin{abstract}
We introduce HCLM, a hierarchical framework for general-purpose cooperative loco-manipulation with dual quadrupedal systems. Coordinating multi-robot collaborative manipulation across floating bases is highly challenging due to the conflicting demands of spatial coordination, robust locomotion, and closed-chain physical interactions. To resolve this, our architecture systematically decouples high-level collaborative reasoning from low-level robust motion execution. At the high level, a centralized Joint Diffusion Policy leverages an $SE(3)$-invariant task-space representation to learn coordinate-agnostic spatial coordination patterns. To translate these frame-agnostic references into physical motion, a task-centric hybrid Whole-Body Controller synergizes a \textit{proactive} kinematic Model Predictive Control for collision-free velocity distribution with a \textit{reactive} execution layer. Crucially, this reactive layer guarantees rapid responsiveness for precise end-effector tracking, while concurrently integrating active force regulation via a cooperative admittance scheme to safely resolve kinematic conflicts and strictly regulate internal stresses during closed-chain interactions. We validate the framework across progressively challenging simulated scenarios, including cooperative carrying, packing and handovers, and successfully deploy the latter in the real world. The results demonstrate reliable task execution, strict configuration agnosticism, and exceptional resilience against severe physical perturbations, offering a highly robust pathway for multi-robot embodied coordination.
\end{abstract}



\section{Introduction}
\label{sec:Introduction}

Real-world mobile manipulation tasks require robotic systems to integrate stable locomotion with precise object interaction. Although single-agent manipulation has advanced significantly, its limited kinematic reach inherently restricts the capacity for spatially distributed interactions and multi-point coordination.  To overcome these limitations, collaborative quadrupedal manipulators offer dynamically reconfigurable workspaces and expanded operational ranges, enabling complex, spatially coordinated tasks. Nevertheless, realizing robust dual-robot loco-manipulation presents two cooperative challenges: first, at the high level, learning coordinate-free policies that enable spatially separated agents to execute diverse manipulation skills rather than pre-defined actions across varied configurations; and second, at the low level, executing stable whole-body control to guarantee precise task-space tracking, reject floating-base disturbances, and strictly regulate internal forces within closed kinematic chains.

Current approaches to multi-robot manipulation systems often fall short of meeting these requirements. Research is typically confined to cooperative locomotion without manipulation \cite{kim2023layered}, or limited to non-prehensile interactions (e.g., pushing) \cite{feng2025learning}, which inherently constrains task complexity. Although cooperative transportation with mobile manipulators has been achieved, these frameworks are often designed for specific, pre-defined tasks \cite{an2025collaborative}\cite{pandit2025multi}, lacking the scalability in task complexity to extend to new scenarios or achieve fine-grained 6D pose control. Consequently, a framework for precise, safe, and general-purpose collaborative manipulation across multiple floating bases is still lacking. To address this, we extend Imitation Learning (IL)—highly successful in single-agent settings \cite{ze20243d}\cite{ze2025generalizable}—into our multi-agent framework, empowering quadrupedal teams to learn diverse collaborative skills.

\begin{figure}[t]
    \centering
    \includegraphics[width=8.6cm]{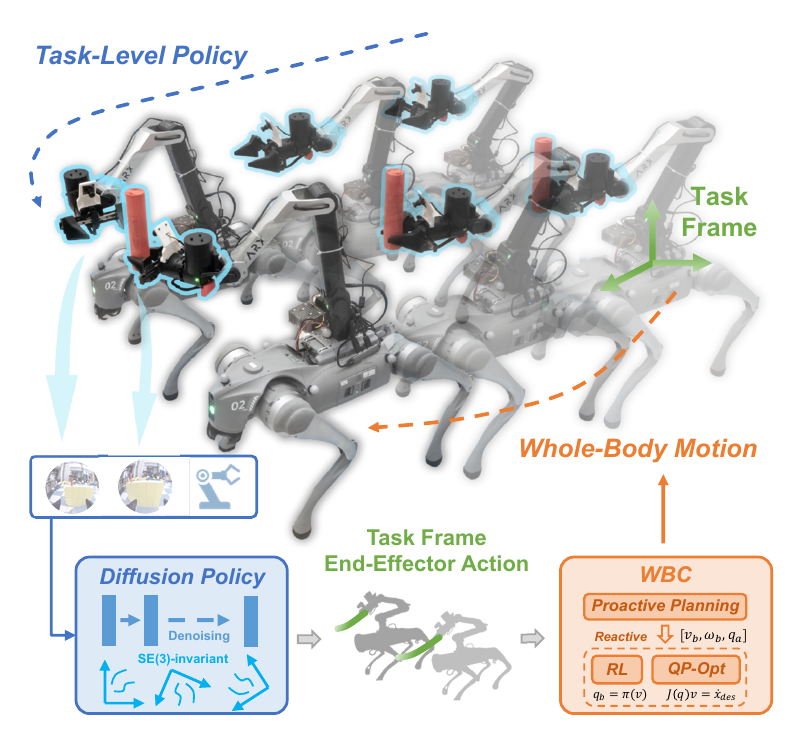}
    \caption{The HCLM Framework. Our hierarchical system enables dual quadrupedal manipulators to perform complex, general-purpose cooperative loco-manipulation tasks in the real world, seamlessly integrating high-level imitation learning with robust low-level whole-body control.}
    \label{fig:overview}
    \vspace{-0.5cm}
\end{figure}

We propose HCLM, a hierarchical framework for general-purpose cooperative loco-manipulation designed for dual quadrupedal systems. To bridge the gap between complex spatial coordination and high-frequency dynamic control, our framework decouples the system into a high-level Diffusion Policy and a low-level task-centric Whole-Body Controller (WBC). To ensure coordination policies generalize across diverse initial configurations, the high-level policy leverages Imitation Learning to generate synchronized end-effector trajectories within an $SE(3)$-invariant task-space. To simultaneously achieve robust, safe locomotion and precise, rapid end-effector tracking, the hybrid WBC synergizes proactive planning with reactive stabilization. Specifically, a kinematic Model Predictive Control (MPC) proactively computes optimal, collision-free motion distributions. Concurrently, a high-frequency reactive layer employs an RL policy for robust base locomotion and a Quadratic Programming (QP) solver for instantaneous manipulator control. By explicitly integrating grasp-matrix-based internal force constraints into the QP, this hierarchical design prioritizes precise end-effector tracking, actively rejecting disturbances while regulating closed-chain interaction forces.

We validated our framework across three challenging simulated tasks: Dynamic Handover, Cooperative Carrying, and Cooperative Packing, and successfully deployed the Dynamic Handover on real-world hardware. The results demonstrate reliable collaboration with high success rates. Crucially, the system exhibits exceptional robustness against severe external physical perturbations and kinematic inconsistencies induced by state estimation drift.

The main contributions of this work are summarized as follows:
\begin{enumerate}
    \item We propose HCLM, a hierarchical framework for general-purpose cooperative loco-manipulation. By systematically decoupling high-level spatial coordination from low-level dynamic execution, it enables dual quadrupedal manipulators to seamlessly integrate stable locomotion with precise object interaction for complex, spatially distributed tasks.
    
    \item We introduce an $SE(3)$-invariant task-space representation to learn frame-agnostic cooperative policies, which are then robustly executed by a hybrid, task-centric WBC. By synergizing proactive motion distribution with reactive whole-body stabilization, this low-level controller guarantees precise end-effector tracking while safely resolving closed-chain kinematic conflicts.
    
    \item We validate our framework across three progressively challenging collaborative tasks. We demonstrate the system's reliable coordination and exceptional robustness against physical perturbations through both simulation and real-world hardware deployments.

\end{enumerate}
\section{Related Works}
\label{sec:related_works}

\subsection{Legged Mobile Manipulation}

Recent advancements in loco-manipulation across quad-rupedal and humanoid platforms \cite{gu2025humanoid} generally fall into two dominant paradigms: optimization-based and learning-based control. Purely optimization-based techniques, notably Model Predictive Control (MPC), excel at precise end-effector tracking and inherently facilitate contact force regulation \cite{molnar2025whole}\cite{sleiman2021unified}. However, their performance is notoriously susceptible to unmodeled dynamics and severe external disturbances\cite{kang2023rl+}. Conversely, learning-based approaches yield highly robust and agile motion generation, yet they typically confine end-effector control to the local body frame \cite{pan2025roboduet}\cite{fu2023deep}, thereby lacking the global task-space coordination required for complex manipulation. To synergize these strengths, hybrid architectures \cite{ma2022combining} deploy RL for resilient base locomotion alongside MPC for precise manipulator control. While recent frameworks like UMI-on-Legs \cite{ha2024umi} attempt to address whole-body task-space tracking, their operational reach remains constrained, struggling with effective motion distribution over long-horizon tasks. 

\subsection{Multi-Robot Collaborative Manipulation}

Current studies on multi-robot collaboration are often fragmented into isolated sub-fields. Some works concentrate solely on cooperative locomotion \cite{kim2023layered}\cite{de2023centralized}, devoid of manipulation skills, while others explore non-prehensile manipulation \cite{feng2025learning} by leveraging structural parts (e.g., legs or heads) for tasks like pushing. Although some approaches incorporate mobile manipulators, they are predominantly limited to pre-defined cooperative transportation scenarios \cite{pandit2025multi}\cite{an2025collaborative}\cite{turrisi2024pacc}. Consequently, these methods lack the scalability required for complex, general-purpose manipulation and fail to realize fine-grained 6D pose control.

Beyond kinematic coordination, closed-chain physical interactions demand rigorous dynamic formulations. Classical approaches utilize grasp matrices to orthogonally decompose contact wrenches into motion-inducing external forces and object-squeezing internal stresses \cite{khatib1996coordination}\cite{erhart2015internal}, actively regulating the latter via impedance or admittance control \cite{de2022decoupling}. Building on these foundations, we embed a cooperative admittance scheme into the null-space optimization. This explicitly regulates internal forces, safely bridging the gap between purely data-driven kinematic policies and the strict dynamic constraints of closed-chain manipulation.

\subsection{Imitation Learning for Mobile Manipulation}

Imitation Learning (IL) has emerged as a highly effective paradigm for general robotic manipulation. However, conventional IL frameworks are frequently tethered to fixed third-person cameras or global point clouds, fundamentally limiting their applicability to mobile platforms \cite{chi2025diffusion, ze20243d}. While recent advancements have addressed mobility by conditioning policies on egocentric visual observations \cite{ze2025generalizable}, the application of IL to multi-robot cooperative manipulation remains surprisingly nascent. Although existing hierarchical architectures \cite{ha2024umi, gupta2025umi} successfully integrate high-level IL policies with low-level whole-body controllers (e.g., MPC or RL) for precise end-effector tracking, they are inherently engineered for single-agent execution. Consequently, these approaches are designed for single-agent strategy execution and lack the coordination mechanisms required for multi-robot systems.
\begin{figure*}[ht]
    \centering
    \includegraphics[width=1\linewidth]{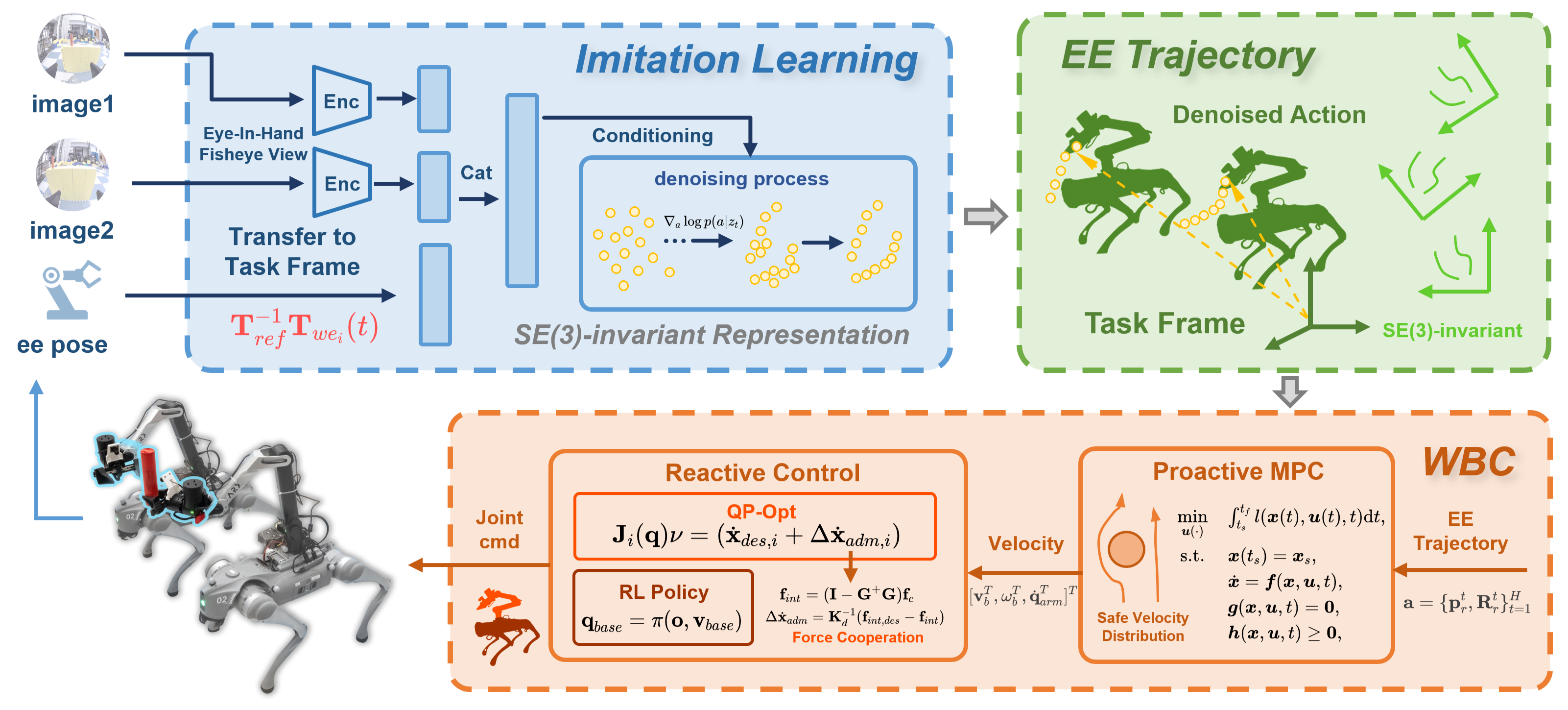}
    \caption{\textbf{Overview of the HCLM framework.} The high-level Imitation Learning module leverages an $SE(3)$-invariant representation to generate synchronized, frame-agnostic end-effector trajectories via a joint diffusion process. To translate these commands into physical execution, the task-centric Whole-Body Controller (WBC) deploys a proactive kinematic MPC for collision-free velocity distribution. Concurrently, a reactive hybrid layer ensures robust execution: an RL policy stabilizes base locomotion, while a QP solver computes arm commands and actively regulates closed-chain internal forces.}
    \label{fig:framework}
    \vspace{-0.3cm}
\end{figure*}

\section{Methods}
\label{sec:methods}

In this section, we present our hierarchical learning framework for dual quadruped-manipulator systems. An
overview of the system is provided in Fig.~\ref{fig:framework}.


\subsection{System Design}

At the high level, a centralized Canonicalized Joint Diffusion Policy infers synchronized end-effector (EE) trajectories from eye-in-hand visual and proprioceptive observations. By operating within an $SE(3)$-invariant task space, the policy intrinsically bypasses global pose estimation drift, learning coordinate-agnostic spatial coordination patterns independent of the underlying floating-base dynamics.

To translate these frame-agnostic references into physical motion, a task-centric Whole-Body Controller dynamically distributes commands across the dual-robot system. Within this low level layer, a \textit{proactive} kinematic Model Predictive Control (MPC) optimizes base velocities and arm postures over a receding horizon to ensure collision avoidance. Concurrently, a \textit{reactive} hybrid execution layer takes over: a dedicated Reinforcement Learning (RL) policy robustly tracks the base locomotion, while a Quadratic Programming (QP) solver computes instantaneous arm commands—utilizing null-space optimization and active force regulation to reject dynamic disturbances and safely manage closed-chain interactions.

Due to the computational demands of the generative model, the Diffusion Policy operates asynchronously on a remote server, whereas the hybrid WBC runs strictly onboard at a 100 Hz control frequency. The following subsections detail each hierarchical component.

\subsection{Canonicalized Joint Diffusion Policy}

We formulate dual-agent collaboration as a centralized imitation learning problem. To achieve implicit intent sharing and tight coordination, we concatenate the state histories of both agents\cite{zhao2023learning} to train a joint Diffusion Policy $\pi_\theta: \mathcal{O} \rightarrow \mathcal{A}$. At each time step $t$, the policy processes an observation history $\mathbf{O}_t = \{\mathbf{o}_{t-T_o+1}, \dots, \mathbf{o}_t\}$ to predict a future action sequence $\mathbf{A}_t = \{\mathbf{a}_t, \dots, \mathbf{a}_{t+T_p-1}\}$. Following a receding horizon paradigm, only the first $T_a$ actions are executed closed-loop to maintain system reactivity.


Although computationally efficient, centralized policies parameterized in absolute global coordinates fundamentally suffer from poor generalization to novel configurations and high susceptibility to state estimation drift. To resolve this, we structure our multi-modal observations—comprising eye-in-hand fisheye images and end-effector poses—within an $SE(3)$-invariant task-space. This strictly decouples the joint policy from any absolute reference frame. By relying exclusively on the relative poses between the two robots, this representation is fundamentally better suited for capturing their tightly coupled coordination patterns.

To implement this canonicalized relative observation space, we define a canonical task reference frame, denoted as $\mathbf{T}_{ref}$. Let $\mathbf{T}_{wb_i}(t) \in SE(3)$ and $\mathbf{T}_{we_i}(t) \in SE(3)$ denote the poses of the $i$-th robot's base and end-effector in the world frame at time $t$, respectively ($i \in \{1, 2\}$). We anchor $\mathbf{T}_{ref}$ to the base pose of the primary robot (Robot 1) at the initial time step $t=0$: $\mathbf{T}_{ref} = \mathbf{T}_{wb_1}(0)$.


Consequently, all input EE poses are transformed into this relative reference frame as $\mathbf{T}_{ref}^{-1}\mathbf{T}_{we_i}(t)$. By unifying the state representations of both agents within the primary robot's local frame, the network is forced to explicitly capture the inter-agent relative spatial constraints—a fundamental requirement for tightly coupled tasks like dynamic handovers. For the action space, rather than predicting incremental delta poses, we parameterize the action $\mathbf{a}_t$ as absolute target poses (a 3D position and 4D quaternion, concatenated with a binary gripper state) defined strictly within $\mathbf{T}_{ref}$ to ensure global trajectory consistency. This formulation mathematically ensures spatial invariance: applying an arbitrary global rigid body transformation $g \in SE(3)$ to the entire system leaves the relative inputs and outputs unchanged. Specifically, the transformed observation under $g$ becomes:
$$ (g\mathbf{T}_{ref})^{-1}(g\mathbf{T}_{we_i}) = \mathbf{T}_{ref}^{-1}g^{-1}g\mathbf{T}_{we_i} = \mathbf{T}_{ref}^{-1}\mathbf{T}_{we_i}, $$
precisely preserving the underlying relative geometric structure without the need for specialized equivariant network architectures \cite{wang2025practical}.

Leveraging this canonicalized representation, $\pi_\theta$ is trained on an expert dataset $\mathcal{D}_{expert}$ comprising synchronized camera streams, end-effector poses, and gripper states. Crucially, this base-centric formulation enhances data extensibility, allowing seamless integration of human demonstrations (e.g., via UMI \cite{liu2024fastumi}) by directly projecting the raw poses into $\mathbf{T}_{ref}$. During preprocessing, all kinematic trajectories are transformed accordingly and low-pass filtered to mitigate measurement noise.

\subsection{Task-Centric Cooperative Whole-Body Control}

The Canonicalized Joint Diffusion Policy generates synchronized task-space end-effector (EE) trajectories, parameterized as $\mathbf{a} = \{\mathbf{p}_{r}^t, \mathbf{R}_{r}^t\}_{t=1}^H\in SE(3)^H$. Executing these across dual floating bases is highly challenging: it requires reconciling the slow, underactuated base dynamics with high-bandwidth manipulation, while strictly regulating closed-chain internal forces. To resolve this, we propose a hybrid Whole-Body Controller that decouples execution into two layers. A proactive kinematic MPC computes optimal, collision-free velocity distributions. Concurrently, a reactive high-frequency layer—synergizing Reinforcement Learning for base stabilization and Quadratic Programming for instantaneous arm control—guarantees rapid EE tracking, dynamic disturbance rejection, and active force compliance.

\subsubsection{Proactive Layer: Kinematic Model Predictive Control}
We employ a kinematic MPC as the proactive layer to resolve long-horizon end-effector tracking and dynamically distribute velocities across the system. By anticipating future states, this receding-horizon formulation naturally integrates real-time obstacle avoidance to ensure operational safety. Modeling the robot as a floating-base manipulator, we define the state vector $\mathbf{x} = [\mathbf{p}_b^T, \mathbf{\Theta}_b^T, \mathbf{q}_{arm}^T]^T$ and control input $\mathbf{u} = [\mathbf{v}_b^T, \boldsymbol{\omega}_b^T, \dot{\mathbf{q}}_{arm}^T]^T$. Here, $\mathbf{p}_b, \mathbf{\Theta}_b \in \mathbb{R}^3$ denote the position and Euler orientation of the base, and $\mathbf{q}_{arm}$ represents the manipulator joint angles. Correspondingly, $\mathbf{v}_b, \boldsymbol{\omega}_b \in \mathbb{R}^3$ are the base linear and angular velocities, and $\dot{\mathbf{q}}_{arm}$ denotes the arm joint velocities. We formulate the MPC at the kinematic level because solving full nonlinear dynamics for an 18-DoF system onboard an edge device is computationally prohibitive. Consequently, the MPC acts purely as a collision-aware velocity reference generator, explicitly delegating dynamic stabilization and disturbance rejection to the high-frequency reactive layer.

The optimal control problem is formulated as follows:
\begin{subequations}
\begin{align}
    \min_{\mathbf{u}_{0:N-1}}  & \sum_{k=0}^{N-1} \left( \|\mathbf{e}_{task, k}\|_{\mathbf{Q}}^2 + \|\mathbf{u}_k\|_{\mathbf{R}}^2 \right) \\
    \text{s.t.} \ \ & \mathbf{x}_{k+1} = f(\mathbf{x}_k, \mathbf{u}_k), \ k = 0, \dots, N-1 \\
    & \mathbf{u}_{min} \leq \mathbf{u}_k \leq \mathbf{u}_{max}, \ \ \mathbf{q}_{min} \leq \mathbf{q}_{arm, k} \leq \mathbf{q}_{max} \\
    & h(\mathbf{x}_{k+1}) \geq (1 - \gamma) h(\mathbf{x}_k), \ k = 0, \dots, N-1 \label{eq:mpc_cbf}
\end{align}
\end{subequations}
where $\mathbf{e}_{task, k}$ is an augmented error vector aggregating both end-effector tracking and manipulator postural objectives, weighted by $\mathbf{Q}$, alongside the control penalty $\mathbf{R}$. Crucially, constraint \eqref{eq:mpc_cbf} integrates a Discrete Control Barrier Function (DCBF) to enforce rigorous \textit{proactive} collision avoidance. By defining a safe set via $h(\mathbf{x}) \ge 0$ and regulating the boundary approach rate with $\gamma \in (0, 1]$, this formulation mathematically guarantees collision-free trajectories regarding both the partner agent and environmental obstacles. The NLP is solved in real-time using the CasADi framework \cite{andersson2019casadi}.

\begin{figure}[t]
    \centering
    \includegraphics[width=8.6cm]{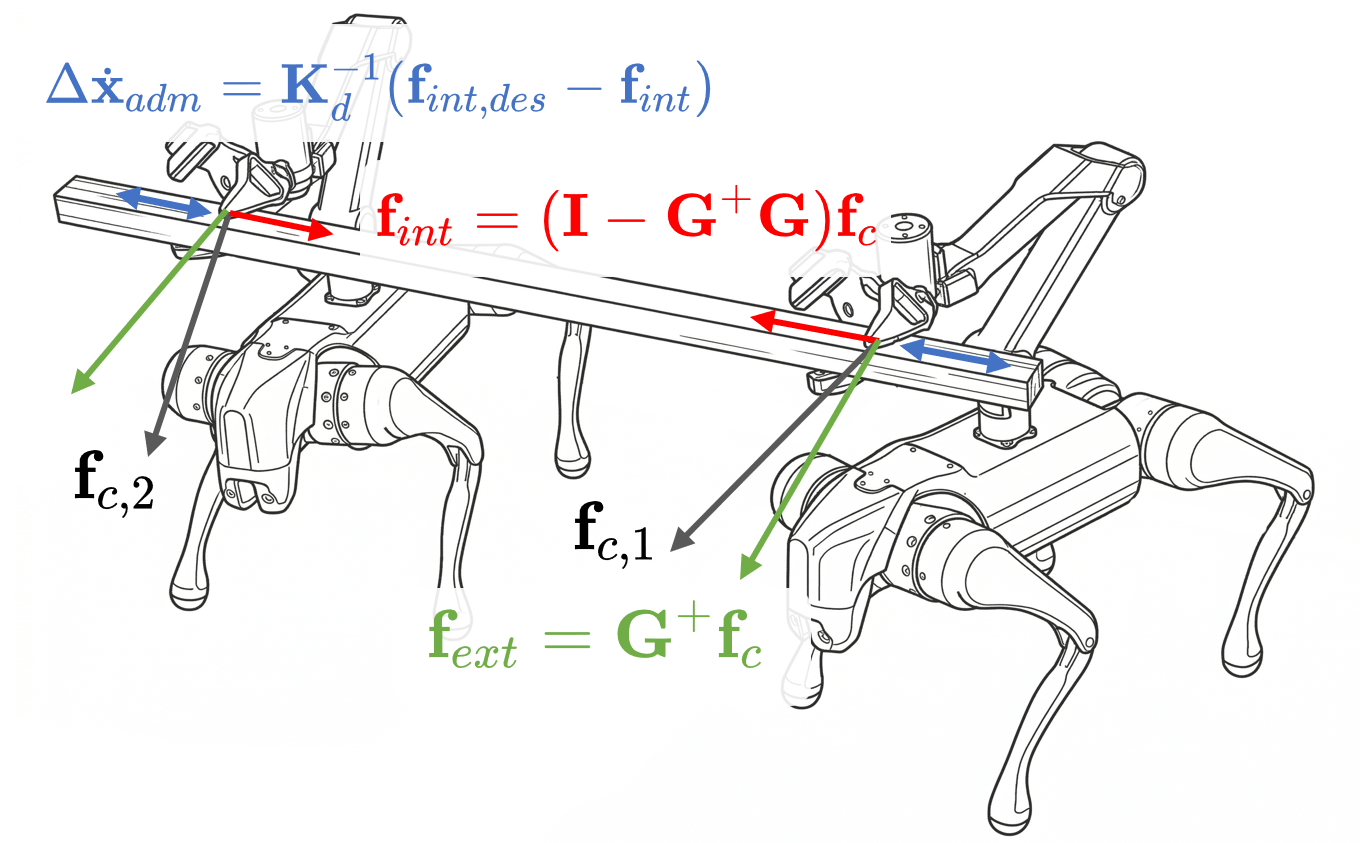}
    \caption{Illustration of the cooperative admittance scheme for closed-chain force regulation. The measured multi-arm contact wrenches ($\mathbf{f}_{c,1}$, $\mathbf{f}_{c,2}$) are orthogonally decomposed using the grasp matrix $\mathbf{G}$. This decoupling systematically isolates the motion-inducing external forces ($\mathbf{f}_{ext}$) from the object-squeezing internal forces ($\mathbf{f}_{int}$) residing in the null space. To safely dissipate closed-chain kinematic conflicts caused by spatial drift, the framework maps the internal force error into a corrective yield velocity ($\Delta \dot{\mathbf{x}}_{adm}$) via a virtual damping matrix $\mathbf{K}_d$, ensuring compliant and stable physical interactions.}
    \label{fig:force}
    \vspace{-0.3cm}
\end{figure}

\subsubsection{Reactive Layer: Hybrid Whole-Body Execution}
While the proactive MPC provides optimal kinematic references, physical execution requires a high-frequency \textit{reactive} layer to compensate for unmodeled multi-body dynamics, payload variations, and external impacts. To achieve this, we decouple the low-level execution: a velocity-level Quadratic Program (QP) computes instantaneous whole-body commands $\boldsymbol{\nu} = [\mathbf{v}_b^T, \boldsymbol{\omega}_b^T, \dot{\mathbf{q}}_{arm}^T]^T$ to reactively balance competing task objectives \cite{pink}, while a dedicated RL policy robustly tracks the resulting base velocities.

At the core of our reactive execution layer is a velocity-level Quadratic Programming (QP) framework that computes instantaneous joint velocities $\boldsymbol{\nu}$. This formulation balances competing whole-body tasks, denoted by the comprehensive set $\mathcal{T}$, which encompasses both end-effector tracking in Cartesian space and nominal posture regulation in joint space. For each task $i \in \mathcal{T}$, $w_i$ dictates its relative priority weight, $\mathbf{J}_i(\mathbf{q})$ represents the analytical task Jacobian, and $\dot{\mathbf{x}}_{des, i}$ denotes the nominal target velocity within its respective task space.

In a standard purely kinematic setup, the QP optimization aims to rigidly track these task-space velocities alongside a regularization term. This baseline formulation is given by:
\begin{subequations}
\begin{align}
    \min_{\boldsymbol{\nu}} \quad & \frac{1}{2} \sum_{i \in \mathcal{T}} w_i \| \mathbf{J}_i(\mathbf{q}) \boldsymbol{\nu} - \dot{\mathbf{x}}_{des, i} \|^2 + \lambda \| \boldsymbol{\nu} \|^2 \\
    \text{s.t.} \quad & \boldsymbol{\nu}_{min} \leq \boldsymbol{\nu} \leq \boldsymbol{\nu}_{max} \label{eq:vel_lim} \\
    & (\mathbf{q}_{min} - \mathbf{q})/\Delta t \leq \boldsymbol{\nu} \leq (\mathbf{q}_{max} - \mathbf{q})/\Delta t \label{eq:pos_lim}
\end{align}
\end{subequations}
where $\lambda$ is a damping factor introduced to mitigate excessive joint velocities near kinematic singularities, and $\Delta t$ is the discrete control step size. 

Crucially, during multi-robot manipulation, rigidly tracking these purely kinematic trajectories generates destructive internal stresses due to inevitable drifts. To safely resolve these closed-chain conflicts, our framework seamlessly integrates a \textit{cooperative admittance scheme} into the QP objective. Let $\mathbf{f}_c = [\mathbf{f}_{c,1}^T, \mathbf{f}_{c,2}^T]^T$ denote the concatenated multi-arm contact wrench. Using a grasp matrix $\mathbf{G}$ \cite{erhart2015internal}, we orthogonally decompose $\mathbf{f}_c$ to isolate the object-squeezing internal forces residing in the null space: $\mathbf{f}_{int} = (\mathbf{I} - \mathbf{G}^+\mathbf{G})\mathbf{f}_c$. We then map this internal force error to a corrective yield velocity $\Delta \dot{\mathbf{x}}_{adm} = \mathbf{K}_d^{-1} (\mathbf{f}_{int, des} - \mathbf{f}_{int})$ for both manipulators, where $\mathbf{K}_d$ is a virtual damping matrix. By superimposing this active compliance term onto the task-space references, the unified QP optimization is reformulated as:
\begin{subequations}
\begin{align}
    \min_{\boldsymbol{\nu}} \quad & \frac{1}{2} \sum_{i \in \mathcal{T}} w_i \| \mathbf{J}_i(\mathbf{q}) \boldsymbol{\nu} - (\dot{\mathbf{x}}_{des, i} + \Delta \dot{\mathbf{x}}_{adm, i}) \|^2 + \lambda \| \boldsymbol{\nu} \|^2 \\
    \text{s.t.} \quad & \text{Constraints (\ref{eq:vel_lim}) and (\ref{eq:pos_lim})}
\end{align}
\end{subequations}
This soft-constrained formulation inherently relaxes rigid positional tracking, actively dissipate closed-chain kinematic conflicts while preserving operational stability\cite{bellicoso2019alma}.

Concurrently, the RL policy ensures the robust execution of the target base velocities $\mathbf{v}_{base}$ commanded by the MPC and refined by the QP. To reject the dynamic coupling and varying interaction forces introduced by the manipulator, the policy is trained using domain randomization—specifically injecting randomized payloads and external wrenches \cite{ma2022combining}. Acting as a highly reactive stabilizer, the policy infers the optimal base joint commands $\mathbf{q}_{base} = \pi(\mathbf{o}, \mathbf{v}_{base})$ directly from proprioceptive observations, maintaining high-fidelity locomotion under complex loco-manipulation scenarios.
\section{Experiments}
\label{sec:experiments}

\begin{figure}[t]
\centering
\includegraphics[width=8.6cm]{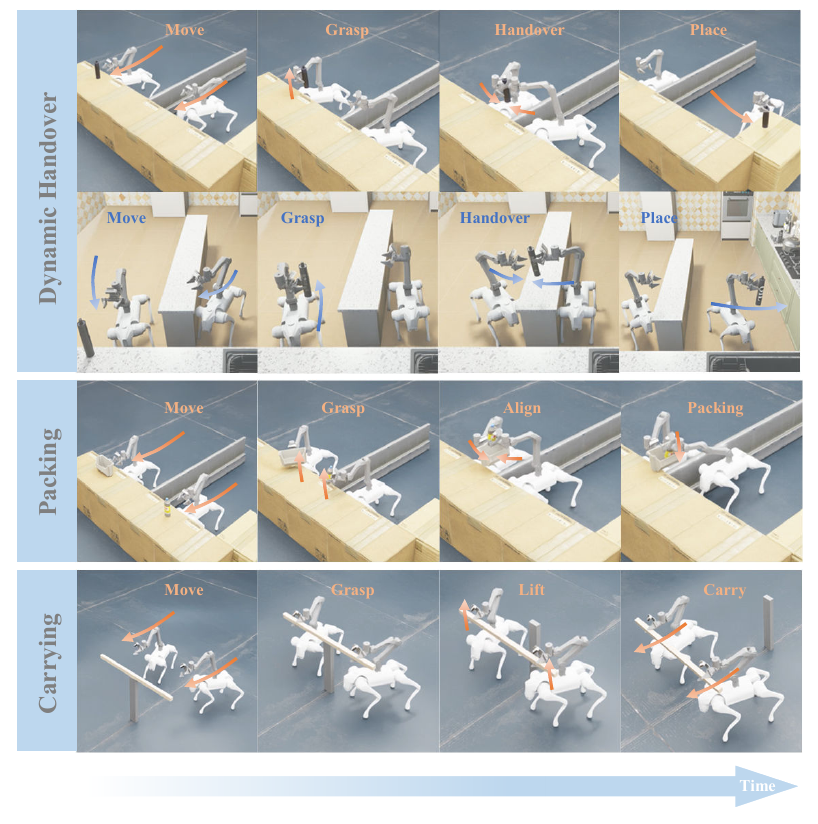}
\caption{Snapshots of the three cooperative loco-manipulation tasks evaluated in simulation: Dynamic Handover, Cooperative Carrying, and Cooperative Packing.}
\label{fig:sim_tasks}
\vspace{-0.3cm}
\end{figure}

To evaluate the performance and robustness of our proposed framework, we conduct extensive experiments across three progressively challenging cooperative loco-manipulation tasks in a high-fidelity simulation environment, as shown in Fig.~\ref{fig:sim_tasks}. Furthermore, we validate the sim-to-real transferability and practical viability of our approach through real-world hardware deployments.

\subsection{Simulation Setup}
\label{sec:sim_setup}

We develop our simulation environment in Isaac Sim, operating the physics engine at 1000 Hz and our unified control framework at 100 Hz. The multi-agent system consists of two quadrupedal manipulators, each combining a 12-DoF Unitree Go2 base with a 6-DoF ARX5 arm. For perception, an eye-in-hand fisheye camera is mounted near the end-effector (EE), providing wide-angle egocentric visual observations.


We test the proposed framework across three distinct scenarios, each highlighting a specific facet of multi-agent coordination:
\begin{itemize}
\item \textbf{Dynamic Handover:} Robot A retrieves a bottle and performs a mid-air transfer to Robot B while both agents remain in continuous motion. Evaluated across two environments (Warehouse and Kitchen), this task demands dynamic spatial targeting and strict inter-agent temporal synchronization. Success requires transferring and placing the bottle at the final target.

\item \textbf{Cooperative Carrying:} The manipulators grasp opposite ends of a rigid rod and collaboratively transport it to a goal region. This scenario evaluates the controllers' ability to maintain closed-chain kinematic constraints during locomotion. Success is defined as navigating to the goal.

\item \textbf{Cooperative Packing:} Robot A lifts and stabilizes a container mid-air while Robot B inserts a bottle into it. This task isolates the system's capacity for high-precision spatial coordination. Success requires accurate insertion without severe collisions or destabilizing the container.
\end{itemize}

\subsection{Simulation Experiments}
Our simulation experiments evaluate the proposed framework across three dimensions: (1) Task Performance: the execution reliability on diverse cooperative loco-manipulation tasks; (2) Configuration Agnosticism: the policy's generalization to arbitrary initial configurations via the $SE(3)$-invariant representation; and (3) Robustness and Compliance: the WBC's capability to reject perturbations and maintaining closed-chain compliance.

\subsubsection{Task Performance on Benchmark Tasks}
\label{sec:benchmark_performance}

To comprehensively evaluate the efficacy of the proposed hierarchical framework, we conduct 20 independent trials for each benchmark task detailed in Section~\ref{sec:sim_setup}. For a rigorous assessment of policy generalization, we compare models parameterized in the absolute world frame (W) against our proposed base-centric canonical frame (BW). Furthermore, to explicitly validate configuration agnosticism, both representations are evaluated under two testing conditions: standard randomized initial setups (Nominal Config) and setups subjected to completely randomized global yaws (Arbitrary Orientations). The quantitative results, measured by overall Success Rate (SR), are summarized in Table~\ref{tab:benchmark_success_rates}.

\begin{table}[t]
\centering
\caption{Success rates across different state representations under nominal and arbitrarily rotated global configurations.}
\label{tab:benchmark_success_rates}
\begin{tabular}{lcccc}
\toprule
\multirow{2}{*}{Task} & \multicolumn{2}{c}{Nominal Config} & \multicolumn{2}{c}{Arbitrary Orientations} \\
\cmidrule(lr){2-3} \cmidrule(lr){4-5}
 & W & BW & W & BW \\
\midrule
Handover (Kitchen)      & 13/20 & 14/20 & 0/20 & 13/20 \\
Handover (Warehouse)    & 12/20 & 13/20 & 0/20 & 14/20 \\
Packing                 & 14/20 & 13/20 & 0/20 & 12/20 \\
Cooperative Carrying    & 13/20 & 13/20 & 0/20 & 11/20 \\
\midrule
\textbf{Total Success}  & \textbf{54/80} & 53/80 & 0/80 & \textbf{50/80} \\
\bottomrule
\end{tabular}

\vspace{-0.3cm}

\end{table}


Table \ref{tab:benchmark_success_rates} summarizes the quantitative performance across the evaluated benchmarks. Under nominal initial configurations, our base-centric formulation (BW) achieves a highly competitive overall success rate of 66.25\% (53/80) across all four dynamic and precision-demanding scenarios. While the absolute world frame baseline (W) performs marginally better in this ideal, unperturbed setting (67.5\% or 54/80), the results clearly validate that our canonicalized relative representation provides highly effective and stable kinematic guidance for complex, dual-agent spatial coordination that is comparable to absolute frame representations.

The critical advantage of our hierarchical framework becomes evident when the system is subjected to arbitrary global orientations. Upon randomizing the initial global yaws, the baseline policy (W) suffers catastrophic failure (0/80), as it fundamentally overfits to the absolute coordinate frame seen during training. In stark contrast, our $SE(3)$-invariant formulation (BW) maintains a highly robust overall success rate of 62.5\% (50/80). This minimal degradation from its nominal performance empirically proves that decoupling the policy from an absolute reference frame successfully achieves configuration agnosticism, ensuring reliable cooperative execution across varied, real-world spatial initializations.

\begin{figure}[t]
\centering
\includegraphics[width=8.6cm]{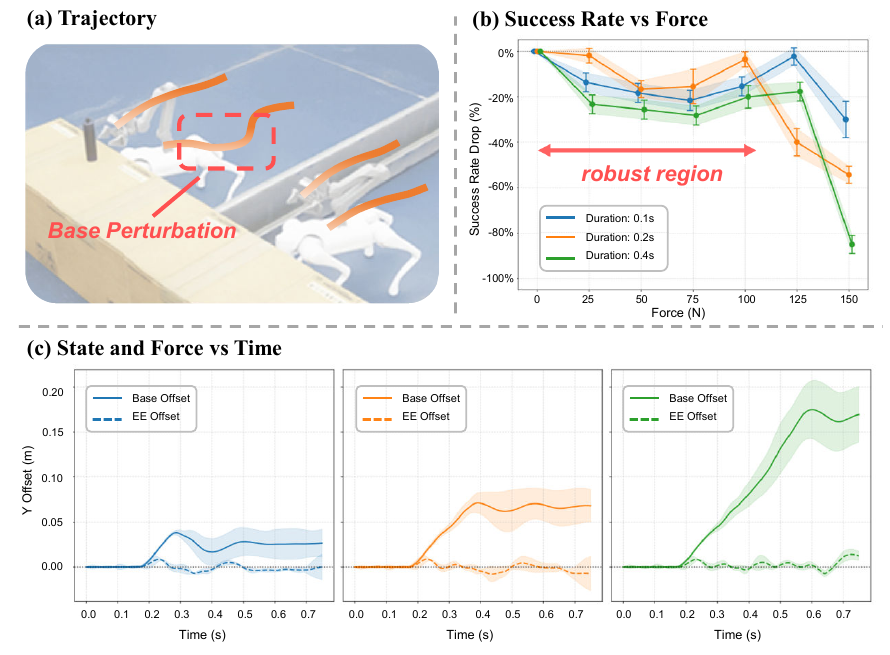}
\caption{System robustness against external base perturbations. (a) Simulated kinematic trajectories illustrating base yielding and EE stabilization. (b) The decrease in task success rates relative to the unperturbed nominal performance, evaluated across varying perturbation magnitudes and durations. (c) Time-series deviation of the base and EE relative to their unperturbed nominal trajectories under a fixed 100 N perturbation with different force durations (blue: 0.1 s, orange: 0.2 s, green: 0.4 s).}
\label{fig:disturbance_rejection}
\vspace{-0.3cm}
\end{figure}

\begin{figure}[t]
    \centering
    \includegraphics[width=8.6cm]{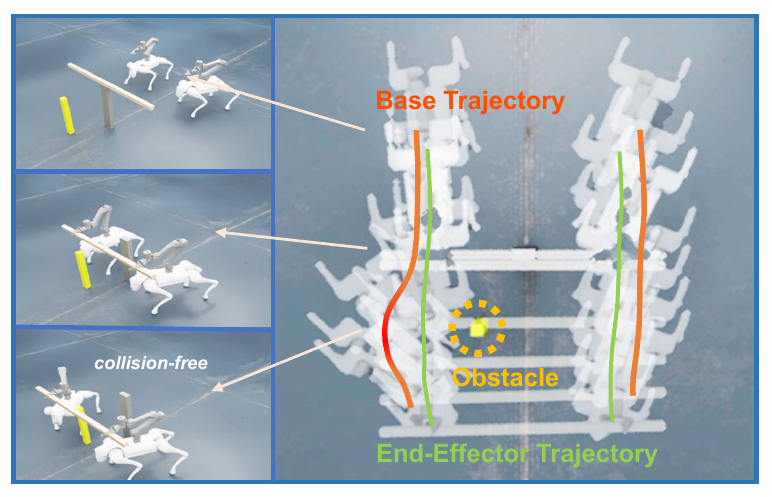}
    \caption{System response to an obstacle. The spatial trajectories illustrate the mobile base autonomously deviating to bypass the threat, while the end-effectors strictly maintain their coordinated collaborative tracking and closed-chain grasp.}
    \label{fig:collision_avoidance}
    \vspace{-0.3cm}
\end{figure}

\subsubsection{Robustness and Capability Analysis}
\label{sec:robustness_analysis}

We systematically investigate the framework's operational resilience against ubiquitous challenges encountered during real-world deployments, specifically evaluating its robustness against external physical perturbations, its capacity for dynamic collision avoidance, and its physical compliance under closed-chain spatial drift.

\paragraph{Robustness to Physical Perturbations on the Base}
To evaluate the disturbance rejection capability of the task-centric WBC, we apply unexpected lateral wrenches to the floating base. As qualitatively depicted in Fig.~\ref{fig:disturbance_rejection}(a), the kinematic trajectories exhibit a distinct spatial decoupling: while the base yields to absorb the external impulse, the end-effector (EE) steadily maintains its target tracking. This behavior is quantitatively verified by the time-series deviation analysis in Fig.~\ref{fig:disturbance_rejection}(c). For instance, under a constant 100~N impact lasting 0.2~s, the EE deviation remains tightly bounded below 2~cm, despite a substantial base displacement of approximately 8~cm. Furthermore, we comprehensively profile the task success rate across varying perturbation magnitudes and durations, as summarized in Fig.~\ref{fig:disturbance_rejection}(b). The results empirically validate the WBC's efficacy in isolating base-level instabilities from critical task-space execution. Notably, the system exhibits a broad ``robust region,'' sustaining highly reliable performance (e.g., restricting the success rate drop to within 10\% to 25\%) even under extreme dynamic perturbations up to 100~N.

\begin{figure}[t]
    \centering
    \includegraphics[width=8.6cm]{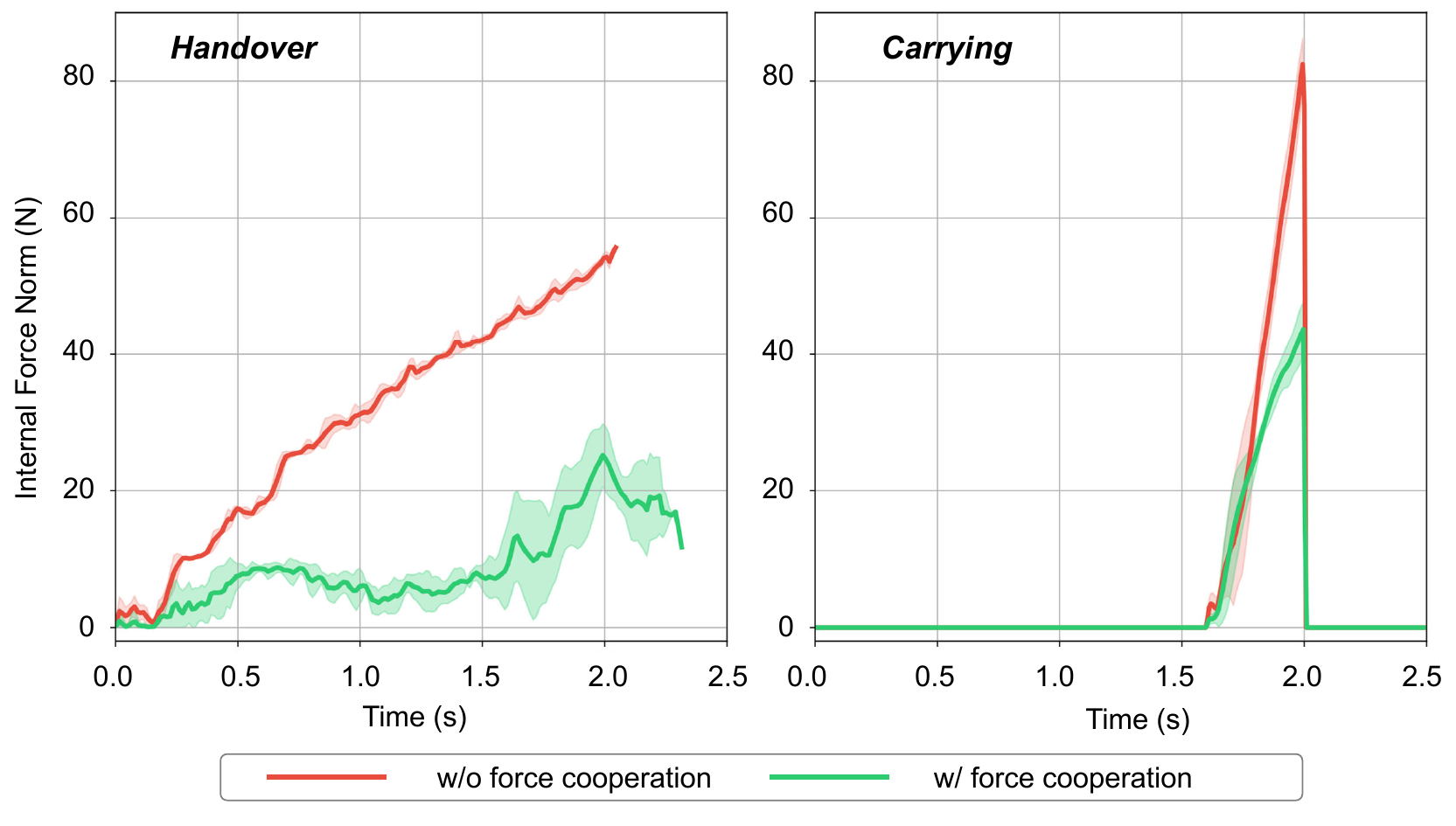}
    \caption{Evaluation of closed-chain compliance and internal force regulation. The temporal evolution of the internal force norm is compared during the Dynamic Handover (left) and Cooperative Carrying (right) tasks, both with (green) and without (red) the proposed cooperative admittance scheme. 
    }
    \label{fig:force_coo}
    \vspace{-0.3cm}
\end{figure}

\paragraph{Dynamic Collision Avoidance}
Finally, we evaluate the system's obstacle avoidance capabilities facilitated by the WBC's null-space optimization. As illustrated by the spatial trajectories in Fig.~\ref{fig:collision_avoidance}, upon encountering a sudden obstacle, the agent autonomously maneuvers its mobile base to bypass the intrusion while dynamically reconfiguring its whole-body posture. Concurrently, the end-effectors strictly track the prescribed task-space trajectory, maintaining a stable grasp on the payload without violating closed-chain constraints. This behavior highlights a fundamental advantage of our hierarchical decoupling: guaranteeing low-level operational safety and reactive compliance without requiring the high-level collaborative policy to replan for unforeseen environmental contingencies.


\paragraph{Closed-Chain Compliance}
To validate the cooperative admittance scheme, we inject uncoordinated spatial drift into the reference trajectories to emulate vision jitter and odometry errors. Under rigid tracking (ablating the admittance module), these deviations induce destructive internal wrenches—exceeding 50~N and 80~N in Handover and Carrying, respectively—causing severe kinematic conflicts and task failure. Conversely, our active admittance scheme maps internal force buildups into corrective yield velocities ($\Delta \dot{\mathbf{x}}_{adm}$). This relaxes rigid positional tracking to dissipate antagonistic forces, successfully preserving closed-chain stability. 

\subsection{Real-World Setups and Experiments}

\begin{figure}[t]
    \centering
    \includegraphics[width=8.6cm]{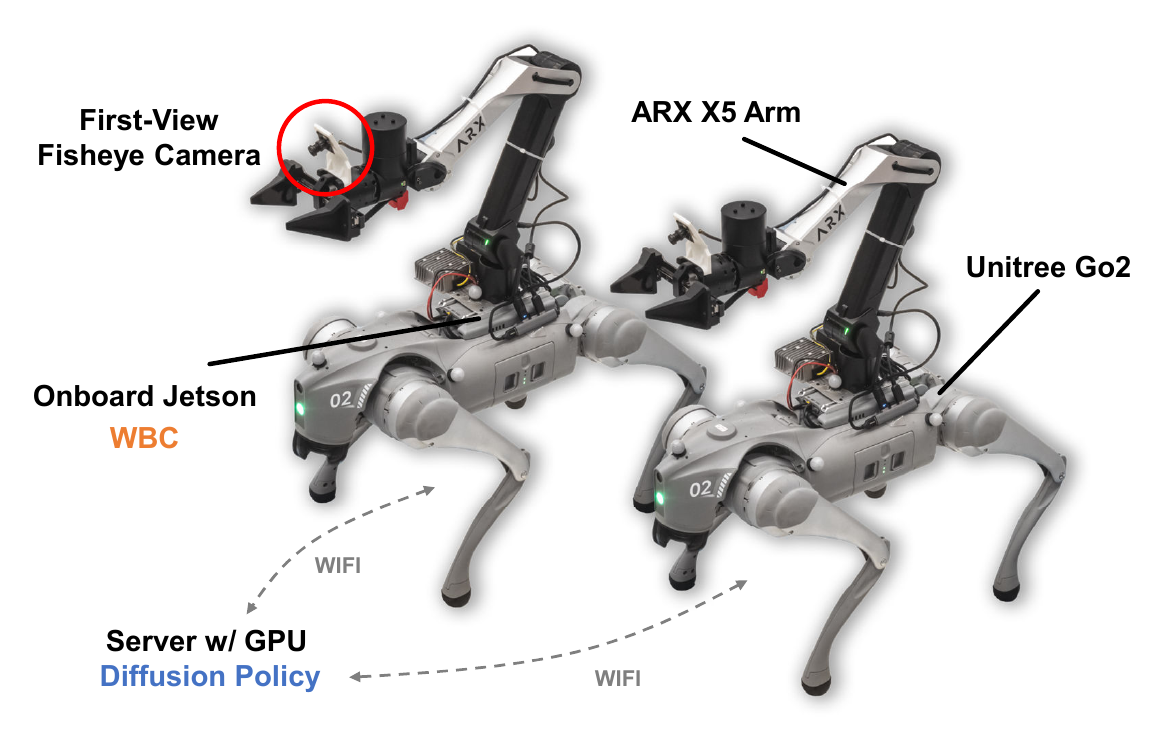}
    \caption{The dual quadruped-manipulator hardware setup for real-world deployment.}
    \label{fig:hardware}
    \vspace{-0.3cm}
\end{figure}

As depicted in Fig.~\ref{fig:hardware}, our fully untethered hardware setup comprises two agents. Each features a 12-DoF Unitree Go2 quadruped, a 6-DoF ARX5 manipulator, and an eye-in-hand fisheye camera. 

Computationally, the low-level Whole-Body Controller executes on the onboard Jetson module at 100~Hz, while the high-level Diffusion Policy infers asynchronously at 20~Hz on a remote desktop (RTX 4060) via Wi-Fi. To ensure rigorous temporal consistency across this distributed architecture, we synchronize the hardware clocks of both agents using \texttt{chronyc}. To address asynchronous communication and multi-rate execution, we implement a two-stage data alignment pipeline. Sensor streams (images, end-effector poses, and joint states) are first locally synchronized on each robot via ROS 1 message filters. These packets are then globally synchronized on the remote desktop, ensuring perfectly time-coherent multi-agent observations for accurate 20~Hz policy inference.


Currently, base poses rely on OptiTrack, with end-effectors computed via forward kinematics. Because our policy conditions on an $SE(3)$-invariant relative representation ($T_{ref}^{-1}T_{we_{i}}(t)$), it remains independent of absolute frames. Thus, while onboard SLAM can seamlessly replace motion capture without policy retraining, future unstructured deployments will require collaborative perception to bound estimation drifts and ensure accurate relative states.

\begin{figure}[t]
    \centering
    \includegraphics[width=8.6cm]{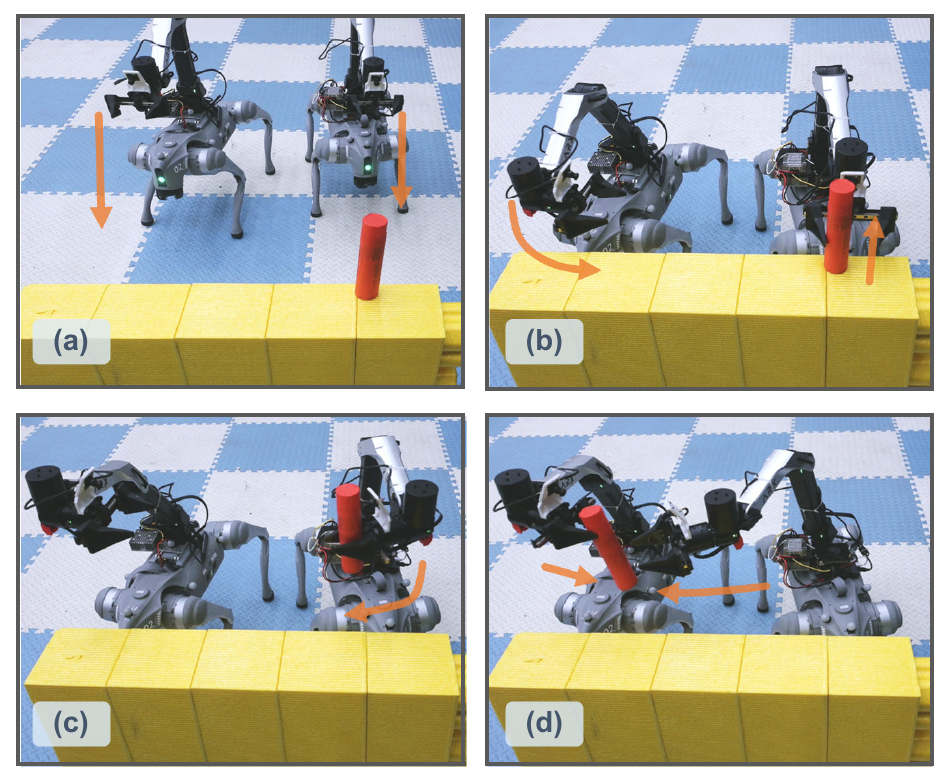}
    \caption{Snapshots of the handover task successfully deployed on the real-world dual quadrupedal system.}
    \label{fig:realexp}
    \vspace{-0.5cm}
\end{figure}

To evaluate sim-to-real transfer and system robustness, we deploy the framework on a real-world dynamic Move-Handover task (Fig.~\ref{fig:realexp}). Due to current hardware sensing constraints, this physical deployment focuses on kinematic coordination and bypasses the closed-chain force regulation. To train the high-level collaborative policy for this specific scenario, we collected a dataset comprising 50 expert demonstrations to capture the necessary spatiotemporal coordination patterns. Operating from randomized initial positions, Robot A retrieves and transfers an object to Robot B while both platforms are in continuous motion. The policy effectively managed relative closing speeds by generating synchronized waypoints, proving that the simulated $SE(3)$-invariant representation successfully generalizes to physical distributed systems.
\section{Conclusions and Limitations}
\label{sec:conclusion}

\textbf{Conclusions.} In this paper, we have presented HCLM, a hierarchical framework for general-purpose cooperative loco-manipulation with dual quadrupedal systems. We have proposed a centralized Joint Diffusion Policy leveraging an $SE(3)$-invariant representation that ensures coordinate-agnostic spatial coordination. Additionally, we have designed a task-centric hybrid Whole-Body Controller incorporating a cooperative admittance scheme to guarantee rapid end-effector tracking, dynamic disturbance rejection, and closed-chain physical compliance. We have conducted extensive simulations and real-world experiments across progressively challenging tasks to demonstrate the effectiveness of our method. Compared to baseline policies parameterized in the absolute world frame, our framework avoids catastrophic failures under arbitrarily rotated global configurations by maintaining a robust 62.5\% success rate, and sustains reliable execution under perturbations up to 100~N.

\textbf{Limitations and Future Work.} First, current hardware deployments rely on motion capture; integrating onboard SLAM is essential for untethered autonomy. Second, expanding our task-specific diffusion policy with Vision-Language-Action (VLA) models would enable semantic reasoning and zero-shot open-vocabulary cooperation. Finally, embedding explicit spatial and dynamic constraints into the imitation policy would prevent kinematically infeasible outputs, further guaranteeing safety before commands reach the WBC.



{
    \bibliographystyle{IEEEtran}
    \bibliography{IEEEabrv, bib/bibliography}
}

\end{document}